\documentclass{article} 
\usepackage{iclr2024_conference,times}

\usepackage{amsmath,amsfonts,bm}

\def\eqref#1{equation~\ref{#1}}

\def\1{\bm{1}}

\DeclareMathAlphabet{\mathsfit}{\encodingdefault}{\sfdefault}{m}{sl}
\SetMathAlphabet{\mathsfit}{bold}{\encodingdefault}{\sfdefault}{bx}{n}

\usepackage{hyperref}
\usepackage{url}

\usepackage{wrapfig,booktabs}
\usepackage[utf8]{inputenc} 
\usepackage[T1]{fontenc}    
\usepackage{booktabs}       
\usepackage{amsfonts}       
\usepackage{nicefrac}       
\usepackage{microtype}      
\usepackage{xcolor}         

\usepackage{amsmath,amssymb,amsfonts}
\usepackage{algorithmic}
\usepackage{graphicx}
\usepackage{textcomp}
\usepackage{xcolor}
\usepackage{caption}
\usepackage{mwe}
\usepackage{subcaption}

\title{Empowering Autonomous Driving with Large Language Models: A Safety Perspective}

\author{%
Yixuan Wang$^1$\thanks{\texttt{Equal Contribution}. Emails: \{yixuanwang2024, ruochen.jiao\}@u.northwestern.edu} \quad Ruochen Jiao$^{1*}$ \quad Sinong Simon Zhan$^1$ \quad Chengtian Lang$^1$ \\ 
\textbf{Chao Huang}$^2$  \quad
\textbf{Zhaoran Wang}$^1$ \quad \textbf{Zhuoran Yang}$^3$ \quad \textbf{Qi Zhu}$^1$ \\ \\
$^1$Northwestern University, USA \quad $^2$University of Southampton, UK \quad $^3$Yale University, USA\\
}

\newenvironment{myitemize}{\begin{list}{$\bullet$}
{\setlength{\topsep}{1mm}
\setlength{\itemsep}{0.25mm}
\setlength{\parsep}{0.25mm}
\setlength{\itemindent}{0mm}
\setlength{\partopsep}{0mm}
\setlength{\labelwidth}{15mm}
\setlength{\leftmargin}{4mm}}}{\end{list}}

\iclrfinalcopy 
\begin{document}

\maketitle

\begin{abstract}
Autonomous Driving (AD) encounters significant safety hurdles in long-tail unforeseen driving scenarios, largely stemming from the non-interpretability and poor generalization of the deep neural networks within the AD system, particularly in out-of-distribution and uncertain data. To this end, this paper explores the integration of Large Language Models (LLMs) into AD systems, leveraging their robust common-sense knowledge and reasoning abilities. The proposed methodologies employ LLMs as intelligent decision-makers in behavioral planning, augmented with a safety verifier shield for contextual safety learning, for enhancing driving performance and safety. We present two key studies in a simulated environment: an adaptive LLM-conditioned Model Predictive Control (MPC) and an LLM-enabled interactive behavior planning scheme with a state machine. Demonstrating superior performance and safety metrics compared to state-of-the-art approaches, our approach shows the promising potential for using LLMs for autonomous vehicles.
\end{abstract}

\section{Introduction}


The current mainstream of autonomous vehicle (AV) software pipeline consists of key modules: perception~\citep{feng2020deep,man2023bev}, prediction~\citep{nayakanti2023wayformer, jiao2022tae}, planning\citep{liu2023safety}, and control. Deep neural networks (DNNs) have become integral to perception and prediction, with a growing interest in planning and control. However, the black-box nature of DNNs, along with their inherent uncertainties from learning algorithms, presents challenges in ensuring the safety of closed-loop AV systems. These challenges are exacerbated by the generalizability issue of DNNs and the prevalence of long-tail driving scenarios not covered during training and design time~\citep{jiao2023semi,fu2023drive,ding2023survey,jiao2023learning}. 

To this end, researchers and engineers in the AV industry are exploring the potential of Large Language Models (LLMs)~\citep{touvron2023llama, openai2020chatgpt3, devlin2018bert} for their ability for human interaction, adept reasoning capabilities, and comprehensive knowledge, particularly in handling long-tail driving scenarios~\citep{yang2023survey,fu2023drive}. Nevertheless, the practical integration of LLMs into the AV software pipeline for safety purposes remains an open question. Therefore, this paper delves into the application of LLMs in autonomous driving from a safety perspective, highlighting its implementation through a couple of illuminating case studies. 

\begin{figure}
    \centering
    \includegraphics[width=0.82\linewidth]{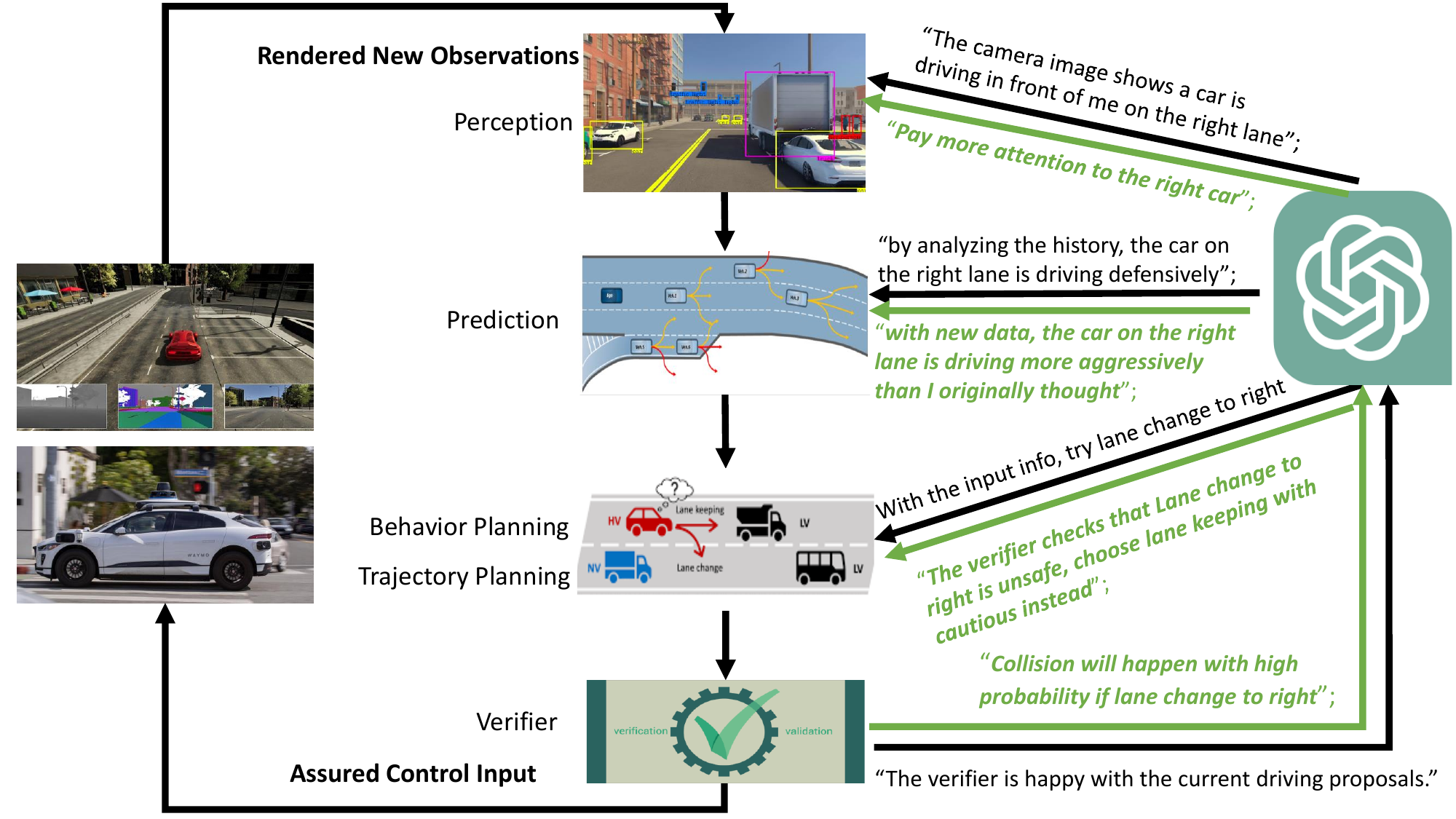}
    \caption{Overview of possible LLM integration for AV with a safety verifier as a shield. Most directly, LLM can make behavior-level decisions such as lane changing by scene understanding via text, which affects the trajectory planning with different safety constraints, as shown in our case studies. The safety verifier checks the safety of the proposed control input from the decision-making and conducts in-context learning if the action is verified to be unsafe, as shown in green arrows. The unsafe feedback can be traced back to the behavior maker, predictor, and perception module as shown. Besides, LLM can assist the perception module in understanding the scene for decision-making better. LLM can also help intention prediction by reading the recent history of the surroundings to better guess their driving habit and intentions (e.g., whether lane changing) for safer decision-making.}
    \label{fig:overview_LLM_AD}
\end{figure}

From a safety perspective, figure~\ref{fig:overview_LLM_AD} shows the possible integration of LLMs for different modules in the AV software pipeline. As a safety-critical system, we equip the AV with a safety verifier for the proposed control input generated from the software stack with assistance from LLMs. The verifier returns safety-checking results to LLM for in-context safety learning which could affect the outputs from different components in various ways.  In this paper, we conduct two case studies to leverage LLM as a behavior-level decision-maker which interacts with a high-level predictor for evaluating the intention and aggressiveness of other agents, and with the low-level trajectory planner and safety verifier. These case studies show that LLM can improve system performance while achieving safety assurance. We hope this paper can provide the AV community with a comprehensive safety standpoint to explore and evaluate the usage of LLM in their AV software stack. 

This paper is organized as follows. We first introduce related works in Section~\ref{sec:related_work}. Section~\ref{sec:case1} and Section~\ref{sec:case2} show our proposed designs integrating LLM as an intelligent safety-aware behavioral decision-maker with a safety verifier and an interactive state machine. Section~\ref{sec:other_roles} discusses the possible integration of LLM for other components including perception, prediction, and simulation in the AV system for safety purposes.  Section~\ref{sec:conclusion} concludes the paper.

\section{Related Works}\label{sec:related_work}
The integration of  LLMs such as GPT-3~\citep{openai2020chatgpt3} into AD has garnered significant attention in recent years, revolutionizing natural language understanding and enhancing the capabilities of self-driving vehicles~\citep{wayve2023lingo}. The related literature from different perspectives is as follows.

\textbf{Human-Oriented}: One direct application is enabling human-vehicle interaction through natural language. LLMs have been leveraged to interpret, respond, and provide suggestions in natural language to human riders and drivers~\citep{zhang2023trafficgpt, wayve2023lingo, xu2023drivegpt4}. These models generate natural language narrations that assist human driving for decision-making and improve the interpretability of AD systems by explaining driving behaviors. Recent works have gone beyond interaction and employed LLMs to learn human driving behaviors and trajectory data through chain-of-thoughts~\citep{wei2022chain, jin2023surrealdriver}. This approach enables the LLM driver to behave like humans to solve complex driving scenarios and even allows LLMs to function directly as motion planners~\citep{mao2023gpt}.  

\textbf{Perception, Prediction, and Planning (Decision-making)}: The reasoning, interpretation, memorization, and decision-making abilities of LLMs contribute to solving long-tail corner cases, improving generalizability, and increasing the interpretability of AD systems. Specifically, there is a growing interest in integrating LLMs into the planning (decision-making) module, which significantly improves user trust and generalizes to various driving cases~\citep{jin2023adapt}. This integration is achieved through fine-tuning pre-trained LLMs~\citep{liu2023mtd} or by prompt engineering with chain-of-thought, which usually enable the AD motion planner to process multilabel inputs, e.g., ego-vehicle information, maps, and perception results~\citep{wen2023dilu, cui2023drive, fu2023drive, mao2023gpt}. Additionally, researchers are exploring LLMs in the perception module to enable self-aware perception, and fast and efficient adaptation to changing driving environments, including tracking, detection, and prediction~\citep{malla2023drama, radford2021learning, wu2023language, ding2023hilm}. Zhou et al summarize the state-of-the-art works in this field~\citep{zhou2023vision}. 

Nevertheless, the aforementioned references fail to address safety concerns associated with LLM in AD. We prioritize safety under the context of LLM, a perspective evident in our case studies. We allow LLM decisions to directly formulate safety constraints for low-level Model Predictive Control (MPC) under prediction uncertainties. Our case studies align closely with the LanguageMPC~\citep{sha2023languagempc}, where the authors also employ LLMs as a decision-maker for AD. They convert LLM decisions into the mathematical representations needed for the low-level controllers, MPC, through guided parameter matrix adaptation. However, LanguageMPC has not been extensively validated in complex driving environments. Additionally, it does not consider uncertainty from predictions nor include safety analysis or optimization in its methodology.

\textbf{Generation and Simulation}: LLMs' generative capabilities have facilitated the acquisition of complex driving data samples, which were previously difficult to gather due to certain environmental constraints. The diffusion model, a method that has recently reached significant success in the text-to-image domain, has become increasingly popular~\citep{sohl2015deep, NEURIPS2020_4c5bcfec}. Some efforts have been put into the area of generating the driving scenarios using diffusion models~\citep{li2023drivingdiffusion, gao2023magicdrive, wang2023drivedreamer, hu2023gaia, zhong2023language}. 

Our work is related to the safety verification for ML-based autonomous systems where AD systems are representative. Safety verification, in general, can be categorized into two groups: 1) explicit reachable set computation~\citep{wang2023polar, huang2022polar, ivanov2021verisig, kochdumper2023provably, goubault2022rino, schilling2022verification, huang2019reachnn} and 2) inexplicit reachable set evaluation, such as barrier certificate~\citep{prajna2006barrier, wang2023joint}, control barrier function~\citep{ames2019control, yang2022differentiable}, forward invariance~\citep{wang2020energy, chen2018data}, etc. There have been emerging works for integrating verification modules into the control learning or reinforcement learning for safety-assured autonomy~\citep{dawson2022safe, wang2023enforcing, wang2023joint, zhan2023state, jin2020neural}. Our paper follows a similar idea where we develop the safety verifier as a shield for the LLM decision-maker to generate safe actions.

\section{LLM Conditioned Adaptive MPC for Trajectory Planning with Safety Assurance}\label{sec:case1}
\begin{figure}
    \centering
\includegraphics[width=0.8\textwidth]{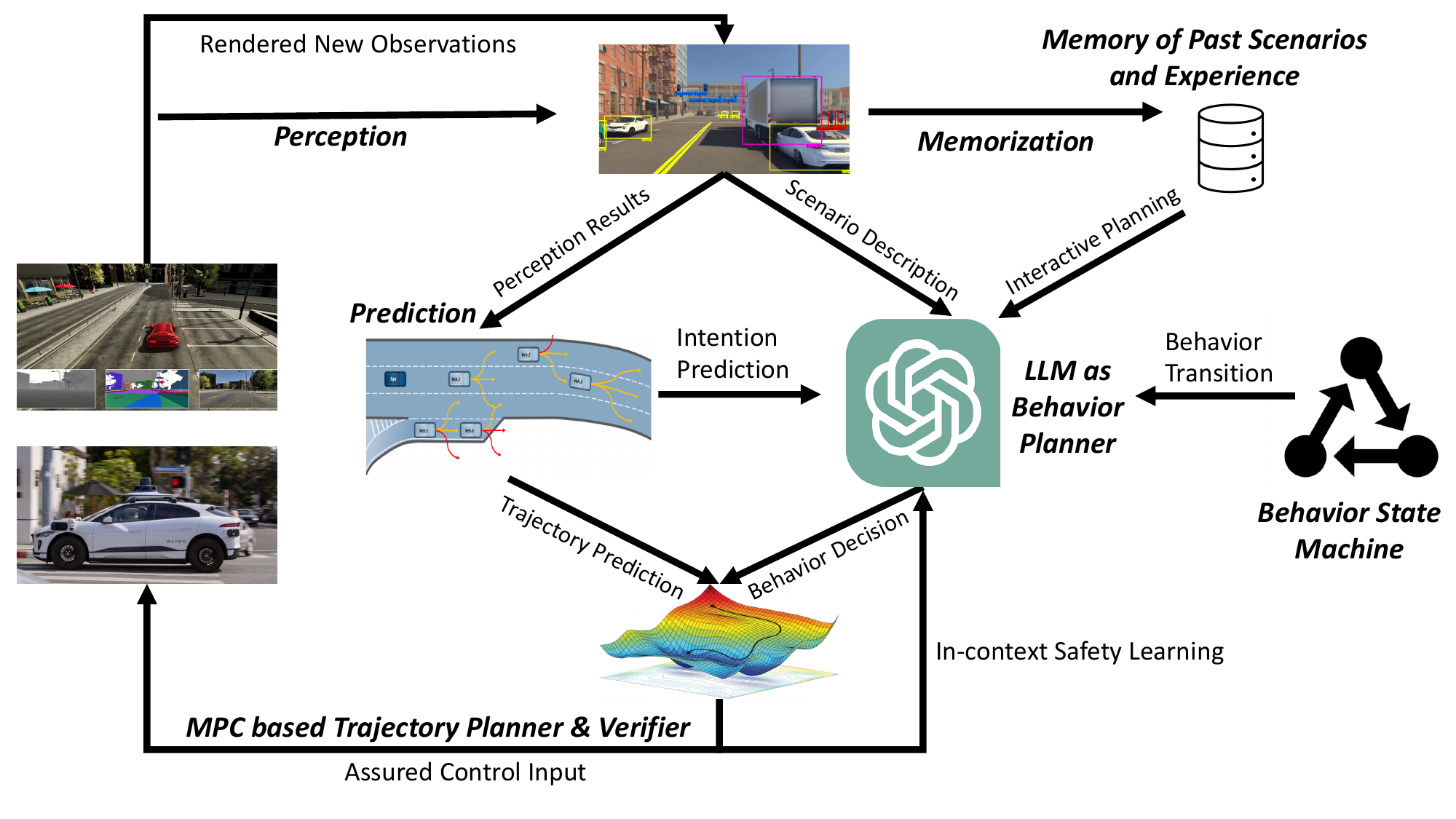}
    \caption{This framework shows LLM as a behavior planner that provides safety constraints for a low-level MPC trajectory planner. The LLM driver takes high-level intention prediction, scenario description, behavior state machine, and its memory via text generated by a template and makes a behavior decision based on its understanding of the driving scene. LLM decisions will formulate safety constraints for low-level MPC-based trajectory planning. Serving as a verifier, the feasibility of the MPC problem will be sent back to LLM to (re)-evaluate its decision for in-context safety learning. }
    \label{fig:llm_behavior_mpc}
\end{figure}

Here we conduct a case study for LLM as a behavior planner via prompt engineering, as shown in Figure~\ref{fig:llm_behavior_mpc}. Next, we introduce the components of this case study as follows.  

\begin{figure*}
        \centering
        \begin{subfigure}[b]{0.475\textwidth}
            \centering
            \includegraphics[width=\textwidth]{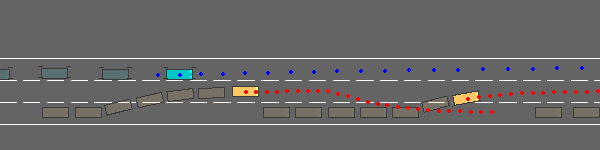}
            \caption[Network2]%
            {{\small \textbf{Lane Keeping:}  The LLM decides to keep the current lane because it is clear and safe. The MPC maintains the highest speed for trajectory planning.}}    
        \end{subfigure}
        \hfill
        \begin{subfigure}[b]{0.475\textwidth}  
            \centering 
            \includegraphics[width=\textwidth]{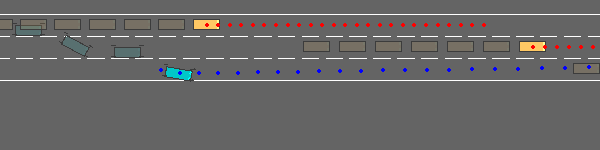}
            \caption[]%
            {{\small \textbf{Lane Change:} LLM decides to change to the rightmost lane from the leftmost lane because the target lane has more space with minimal safety risk.}}    
        \end{subfigure}
        \vskip\baselineskip
        \begin{subfigure}[b]{0.475\textwidth}   
            \centering 
            \includegraphics[width=\textwidth]{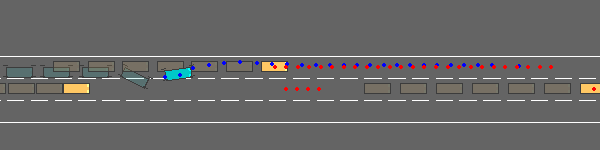}
            \caption[]%
            {{\textbf{Aborted Lane Change:} We discover that LLM can abort its lane-changing if the MPC is infeasible during lane changing to reduce the collision risk.}}    
        \end{subfigure}
        \hfill
        \begin{subfigure}[b]{0.475\textwidth}   
            \centering 
            \includegraphics[width=\textwidth]{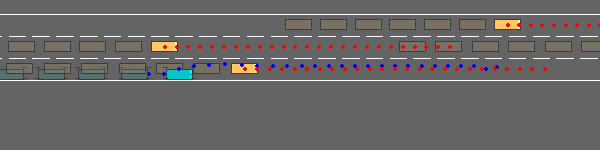}
            \caption[]%
            {{\small \textbf{Failsafe:} The failsafe mode keeps the current lane and maintains a minimal distance for safety. It is only used if LLM's decisions are infeasible in MPC.}} 
        \end{subfigure}
        \caption[ The average and standard deviation of critical parameters ]
        {\small The ego car is in blue and other agents are in yellow. The blue dots are the planned trajectory waypoints of the ego. The red dots are the sampled waypoints of other agents from the interval-based prediction. The grey rectangles are the recent trajectory histories of the ego and other agents. The LLM exhibits safe lane keeping, optimistic lane changing, cautious lane changing abort, and conservative failsafe in the simulations. } 
        \label{fig:highwayenv}
    \end{figure*}

\noindent
\textbf{Environment and System:} Given the safety cost of driving, we primarily focus on a simulated highway-driving environment by using HighwayEnv~\cite{highway-env}. As shown in Figure~\ref{fig:highwayenv}, we consider a one-way three-lane driving scenario. We assume that the vehicle dynamics is known and available to MPC, which can be expressed as $s_{t+1} = f(s_t, u_t)$ where $s = (x, y, v_x, v_y) \in \mathcal{S} \subset \mathbb{R}^4$ with $x, y, v_x, v_y$ denote longitudinal position, lateral position, longitudinal speed, and lateral speed, respectively. The continuous control input to the ego vehicle $u_t \in \mathcal{U} \subset \mathbb{R}^2$ includes acceleration and steering signal. $f: \mathcal{S} \times \mathcal{U} \rightarrow \mathcal{S}$ denotes the bicycle model dynamics~\citep{jiao2023kinematics}.

\noindent
\textbf{Input and Output of LLM:} We call OpenAI GPT-4 API as our LLM driver agent.  We input a template-generated text description of the surroundings within a specific perception range including their relative position (such as "the car $i$ is driving in front of the ego on the right lane" or "the car $i$ is driving behind the ego in the middle lane"), their relative speed (such as "the car $i$ is driving faster/slower than the ego"), the estimation of time to the collision to other agents (relative distance / relative speed), along with other vehicle's intention predictions. The output of the LLM decision maker is constrained to select a target lane for lower level MPC (such as "Middle Lane, Left Lane, Right Lane") with the reasoning. Every decision made by LLM will have 5 consecutive control steps. 

\noindent
\textbf{Prediction Module: } The prediction module on the AV predicts the future state $\hat{s}^j_t$ of surrounding car $j$ at time step $t$. To be realistic and considering uncertainties, we assume the predicted position results are intervals on a specific time step, i.e, instead of $\hat{x}^j_t, \hat{y}^j_t$, we now have $[ \underline{\hat{x}}^j_t, \bar{\hat{x}}^j_t]$ and $[ \underline{\hat{y}}^j_t, \bar{\hat{y}}^j_t]$. We assume the position intervals contain the ground truth $x_t^j, y_t^j$ of the surroundings in the future, i.e., $x_t^j \in [ \underline{\hat{x}}^j_t, \bar{\hat{x}}^j_t]$, $y_t^j \in [ \underline{\hat{y}}^j_t, \bar{\hat{y}}^j_t]$. Because of the receding horizon nature of MPC, we need to call the prediction module to get the prediction results for safety constraint formulation in MPC. 
\noindent
Before introducing our LLM-conditioned MPC, we first show that a \textbf{naive MPC formulation} of trajectory planning as 
\begin{equation}\label{eq:ori_mpc}
    \begin{aligned}
        & \min_{u_t, u_{t+1}, \cdots, u_{t+k}} -x_{t+k} + \sum_{i=t}^{t+k-1} || u _{i+1} - u_{i} ||_2, \\ 
        \text{s.t.},~~ & s_{i+1} = f(s_i, u_i), \forall i \in [t, t+k],\quad
        y_{inf} \leq y_{i} \leq y_{sup}, \forall i \in [t, t+k]\quad \text{(Road boundary)}, \\
        & |x_i - \underline{\hat{x}}^j_i| - L \geq 0,  |x_i - \bar{\hat{x}}^j_i| - L \geq 0, \text{where Lane}([ \underline{\hat{y}}^j_i, \bar{\hat{y}}^j_i]) == \text{Lane}(y_i)\quad \text{(Safety)}
    \end{aligned}
\end{equation}
where $\text{Lane}(y) \in {0, 1, 2}$ is an indicator function that determines which lane the car is driving on by its lateral position $y$, specifically ${0, 1, 2}$ denotes "Left", "Middle", and "Right".  The objective function aims to maximize the performance (longitudinal position or speed) with minimal control jerks. 

\textbf{LLM Conditioned Adaptive MPC for Trajectory Planning:} To reduce the complexity, we leverage the reasoning ability and common sense knowledge of LLM to decide which lane to drive for the MPC, by providing the scene text description to LLM and ask for a decision that relaxes the constraints in MPC. Specifically, at time step $t$, our \textbf{LLM conditioned MPC} tries to solve the following optimization problem
\begin{equation}\label{eq:llm_mpc}
    \begin{aligned}
        & \min_{u_t, u_{t+1}, \cdots, u_{t+k}} -x_{t+k} + \sum_{i=t}^{t+k-1} || u _{i+1} - u_{i} ||_2, \\ 
        \text{s.t.},~& s_{i+1} = f(s_i, u_i), \forall i \in [t, t+k], \quad y_{inf} \leq y_{i} \leq y_{sup}, \forall i \in [t, t+k]\quad \text{(Road boundary)} \\
        & \text{\textbf{Lane}}(y_i) = \text{\textbf{Lane(LLM)}}\quad \text{(\textbf{Behavior provided by LLM})} \\
        & |x_i - \underline{\hat{x}}^j_i| - L \geq 0,  |x_i - \bar{\hat{x}}^j_i| - L \geq 0, \text{where Lane}([ \underline{\hat{y}}^j_i, \bar{\hat{y}}^j_i]) == \text{\textbf{Lane(LLM)}}\quad \text{(Safety)} 
    \end{aligned}
\end{equation}

The problem~\ref{eq:ori_mpc} is harder to solve than problem~\ref{eq:llm_mpc}. The increased complexity originates from the constraint $\text{Lane}([ \underline{\hat{y}}^j_i, \bar{\hat{y}}^j_i]) == \text{Lane}(y_i)$, where $\text{Lane}(y_i)$ is undetermined and can choose from $\{0, 1, 2\}$. Therefore problem~\ref{eq:ori_mpc} is a mixed integer nonlinear programming problem. \textit{In practice, this problem is often infeasible, which is also observed in our case studies.} With the decision from LLM by its knowledge, we remove the integer decision variable in problem~\ref{eq:llm_mpc} and thus it is easier to solve. Our approach shares a similar philosophy of hierarchical MPC as introduced in~\citep{huang2016hierarchical} where we decompose a hard trajectory planning into a two-phase problem that is easier to solve.  

\begin{figure}[h]
    \centering
\includegraphics[width=0.85\linewidth]{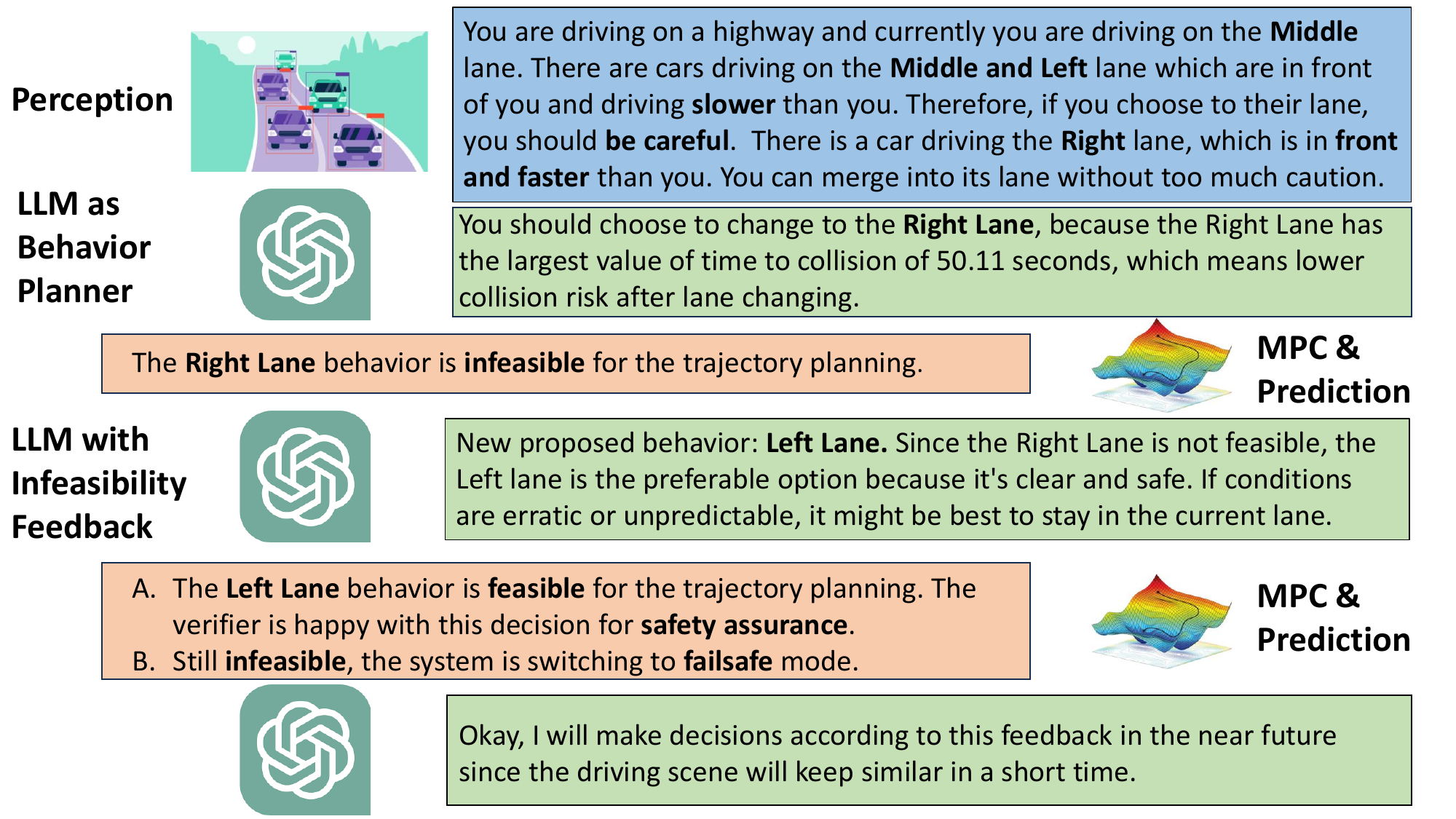}
    \caption{In-context safety learning for LLM with the feedback from MPC for trajectory planning. }
    \label{fig:case1_llm_dialogue}
\end{figure}

\noindent
\textbf{In-context Safety Learning with Verifier: } For safety purposes, control input to the ego vehicle has to go through a  verifier for safety checking and provide the verification result back to the LLM to reevaluate the behavior decision. In general, the verifier could be in the form of reachability analysis~\citep{wang2023polar}, barrier theory, etc~\citep{wang2023enforcing}, as we detailed in the related work. In this case study, we use the \textit{feasibility of the LLM-conditioned MPC~\ref{eq:llm_mpc} as the safety verifier}. If the MPC is feasible which means there exists a safe control signal, we then feedback ``the verifier is happy with the proposed $\textbf{Lane}$'' to LLM. Otherwise, infeasible MPC indicates potential collisions which we feedback to LLM to reevaluate and regenerate another behavior, as shown in Figure~\ref{fig:case1_llm_dialogue}. 

\noindent
\textbf{Failsafe Mode:} It is possible that regenerated behavior or all behaviors are still infeasible for the low-level MPC and thus safety cannot be assured. In this case, we design the AV system switch to a failsafe mode, to keep the current lane and apply a (possibly hard) brake to keep a minimal distance from the front leading car as $-\frac{v_e^2 - v_s^2}{2(\underline{x_l} - x_e - \epsilon)}$ where $v_e, v_s$ are the ego and leading velocity, $\underline{x_l}$ is the lower bound of the estimation for the leading car's location and $x_e$ is the ego position, $\epsilon > 0$. This failsafe optimistically disregards collision with the following car as the ego is optimized to driving faster than the rest IDM-based cars. To be more conservative, one can consider the following car. 

\begin{wraptable}{r}{9.cm}
\centering
  \caption{Comparison results of the case study with 5 episodes.}
  \begin{tabular}{cccl}
    \toprule
      & Safety & Velocity(m/s) & Latency(s)  \\
    \midrule
     Ours & \checkmark & \textbf{34.3}($\pm$7.7) & \textbf{1.7}($\pm$2.7) \\
      \midrule
      DriveLikeAHuman & $\times$ & 31.9($\pm$5.1)  & 55.5($\pm$15.2)  \\
  \bottomrule
\end{tabular}\label{case1_table}
\end{wraptable}
\noindent
\textbf{Experiments Analysis:} We compare our approach with the state-of-the-art open-source DriveLikeAHuman~\citep{fu2023drive} because it also testfies in the HighwayEnv simulator. We add the same interval-based prediction uncertainty to the DriveLikeAHuman framework and adapt its heuristic safety rule considering the interval uncertainty for a fair comparison. We simulate 300 control steps in one test episode. The maximum velocity is set to $40~m/s$. We run 5 trials/episodes for each method and record their results as in Table~\ref{case1_table}. 
\begin{myitemize}
    \item \textit{Safety}: No collision happened in our simulations with 1500 total control steps and more than 300 LLM decision-makings (each decision made by LLM is followed by 5 consecutive control steps). Except for an LLM calling error in one trial, DriveLikeAHuman has collisions in 4 trials around $30\text{th}\sim50\text{th}$ steps. This is because it uses a low-level PID control with a naive high-level heuristic safety rule that does not consider vehicle dynamics and constraints for safety checking.
    \item \textit{Average Velocity}: We measure the longitudinal speed average and standard deviation as performance metrics. The ego drives faster with our approach. This is because we maximize the longitudinal location (speed) in the objective function of our LLM-conditioned MPC. 
    \item \textit{Latency}: The latency of our approach includes the OpenAI API call every 5 control steps and the timing of solving MPC every step while the baseline spends most of the time on the chain-of-thought process with the API per control step. Although both latency are not realistic for real-world driving, ours is significantly shorter than the baseline's. 
\end{myitemize}


\section{LLM as Interactive Decision Maker: Interactive Planning by Behavior Prediction and State Machine  }\label{sec:case2}

\begin{wrapfigure}{r}{0.66\textwidth}
    \centering
    \includegraphics[width=\linewidth]{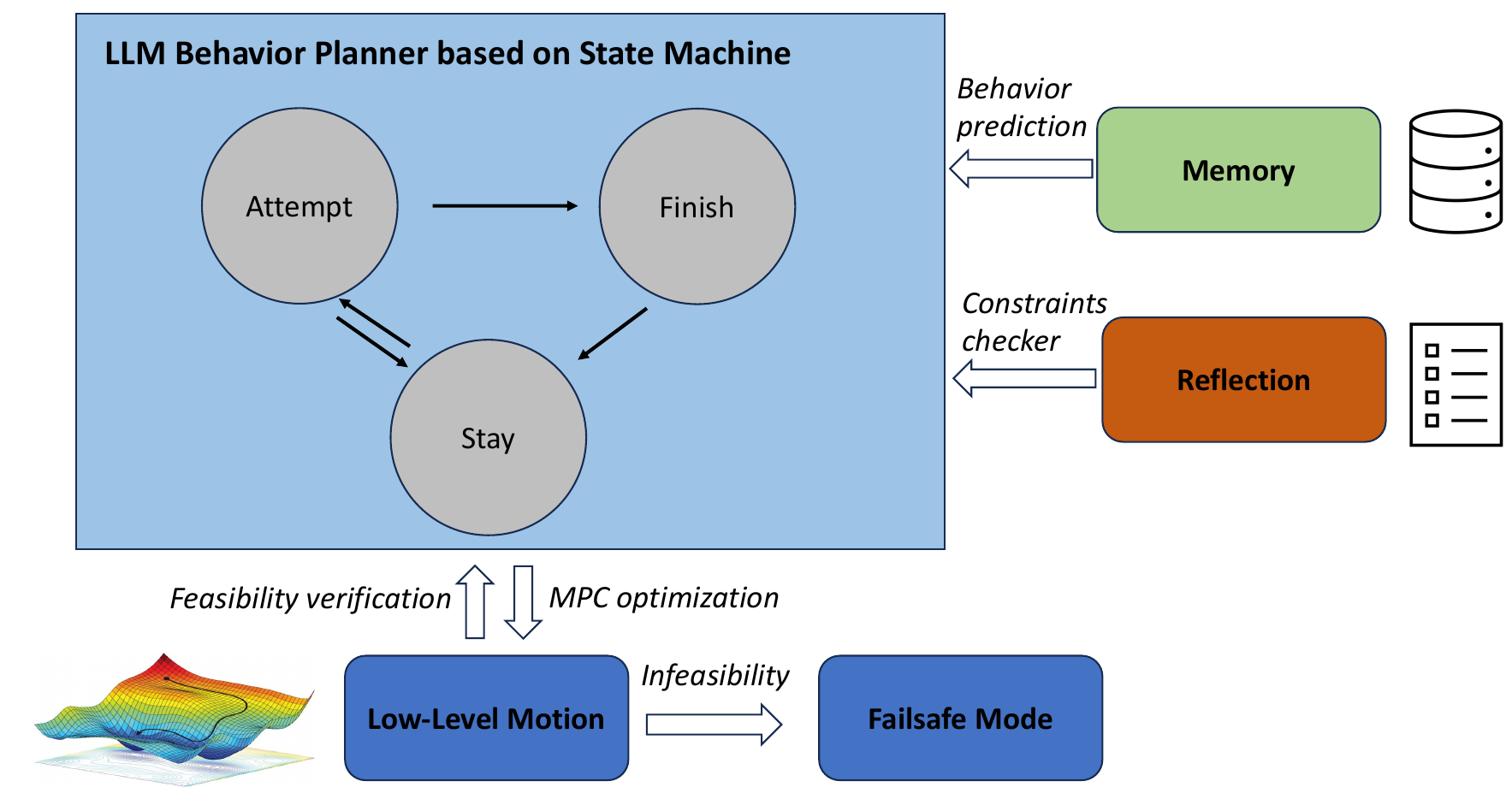}
    \caption{Interactive multi-step decision-making based on the behavior state machine and intention prediction. The memory will store the previous observations of other agents and the reflection module will check hard constraints including state transition rules and minimum safety requirements.}
    \label{fig:case2}
\end{wrapfigure}

As with most existing works on LLM for AD, our previous case study focuses on one-step planning or single-frame decision-making. We can further improve the performance and safety of LLM for driving tasks by explicitly considering the ego vehicle's high-level behavior transitions and the interaction with surrounding agents in multiple consecutive steps. In Figure~\ref{fig:case2}, besides the MPC verifier we proposed previously, we further design the state machine framework as behavior transition guidance, the memory module for intention prediction, and the reflection module for behavior-level safety checks and in-context learning. We will explain them in detail in the following.

\noindent
\textbf{State Machine as Behavior Guidance for LLM Decision Maker: } 
In practical driving situations, executing maneuvers like lane changes or unprotected left turns involves several steps. Vehicles must adhere to established rules or patterns to interact effectively with nearby traffic and accomplish these tasks. This necessitates that  LLMs conform to specific \textit{behavioral constraints}. We propose to integrate a state machine into the LLM through in-context learning. This approach aims to ensure adherence to these essential regulations and established patterns while maintaining sufficient flexibility in decision-making. 

The \textbf{state-machine-conditioned LLM} can enhance the safety and interpretability of the decision-making process in several aspects. First, the state machine can constrain the decision space and simplify the dependency among time-series decisions. Humans can trust the decision-making pipeline if the LLM ensures the transition between states is safe. Second, we add some intermediate/interactive states into the state machine design to help the LLM better understand other vehicles' behavior. During these states, the LLM can proactively interact with the surrounding vehicle (e.g. the following vehicle on the target lane when changing lanes) but still ensure safety. 

\begin{figure}
    \centering
    \includegraphics[width=0.83\linewidth]{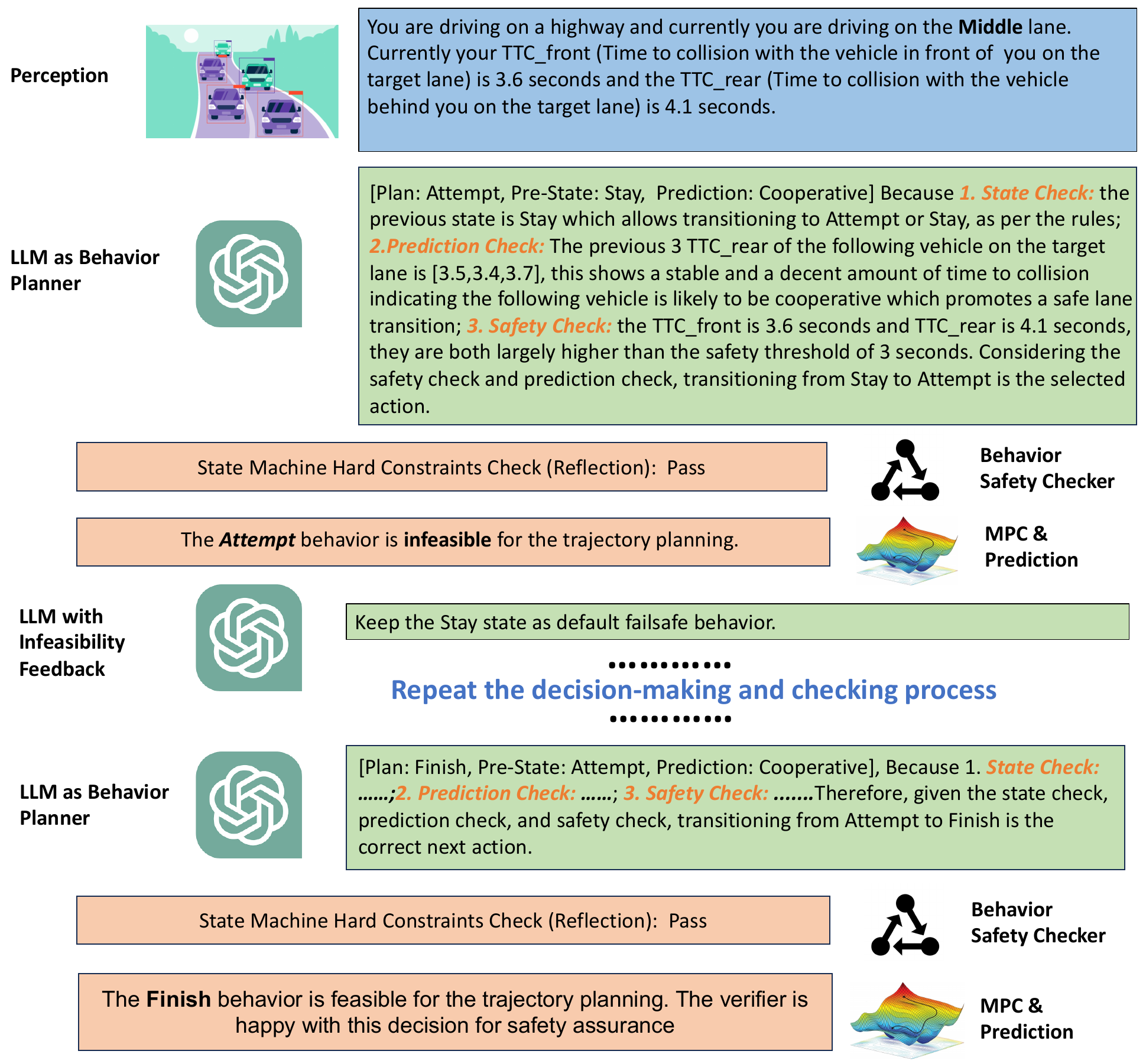}
    \caption{An example of our proposed safe interactive decision-making pipeline for lane changing. With the state machine design and behavior level prediction, the LLM-powered agent can make explainable and safe decisions continually and interactively in complex scenarios. In each cycle, the LLM will reason its decision by three behavior-level checks (state, prediction, and safety). The reflection module will provide feedback for failsafe plans and in-context learning if LLM makes severe and obvious mistakes. The low-level MPC is in charge of the safety verification and execution.}
    \label{fig:case2_llm_dialogue}
\end{figure}

In Figure~\ref{fig:case2}, we present our pipeline for interactive lane changing using LLM as the decision-maker. The framework is centered around the state machine which defines the basic behavior pattern of our LLM. The memory stores important past information about surrounding vehicles, helping the LLM make predictions of their intentions. The reflection module is to monitor the LLM and make sure the transition is valid from state to state and to give feedback to the LLM for in-context learning when the LLM violates hard transition constraints. The LLM determines transitions based on predefined rules and inferred information. The transition involves several checks:

\emph{State Check}: The selected state must be valid as per a predefined state machine graph.

\emph{Safety Check}: The LLM evaluates the possibility of collision if it takes certain actions transiting to the next state. In this particular lane-changing example, the time-to-collision (TTC) is applied to ensure the proposed state won't lead to a collision. The LLM will compare the TTC against a set threshold.

\emph{Prediction Check}: The LLM predicts the intentions of nearby vehicles based on their historical behaviors in past multiple frames in the memory modules. If the LLM deems a surrounding vehicle too aggressive or uncooperative, it's unsafe to proceed with the maneuver. The LLM can interact with the surrounding vehicles in different manners given their different predicted behavior patterns.

 \begin{figure*}
        \centering
        \begin{subfigure}[b]{0.475\textwidth}
            \centering
            \includegraphics[width=\textwidth]{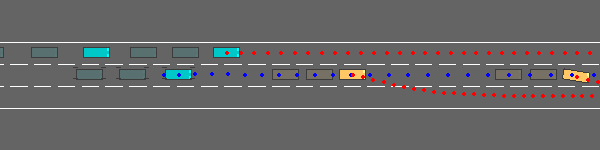}
            \caption[Network2]%
            {{\small \textbf{Phase 1.}  The ego (blue, middle) by LLM aims to cut into the left lane. The LLM  notices there isn't enough space for a safe lane change, picks the "Stay", and accelerates to pass the blue car in front.}}         
        \end{subfigure}
        \hfill
        \begin{subfigure}[b]{0.475\textwidth}  
            \centering 
            \includegraphics[width=\textwidth]{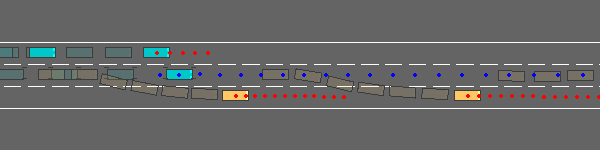}
            \caption[]%
            {{\small \textbf{Phase 2.} Ego vehicle (LLM) is passing the blue car in front and now it only needs to consider and interact with the leading vehicle on the target left lane. LLM decides to continue in the "Stay" in this cycle.}}
        \end{subfigure}
 
        \vskip\baselineskip
        \begin{subfigure}[b]{0.475\textwidth}   
            \centering 
            \includegraphics[width=\textwidth]{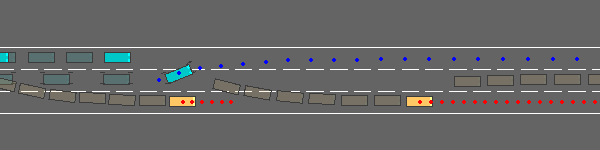}
            \caption[]%
            {{\textbf{Phase 3.} Ego vehicle (LLM) passed the blue car. LLM decides to transit the behavior state to "Attempt" given the comprehensive reasoning including prediction, state transition check and safety analysis. In the state "Attempt", the ego vehicle moves to the middle of two lanes and further observes the reaction of the following vehicle on the target lane. }}    
        \end{subfigure}
        \hfill
        \begin{subfigure}[b]{0.475\textwidth}   
            \centering 
            \includegraphics[width=\textwidth]{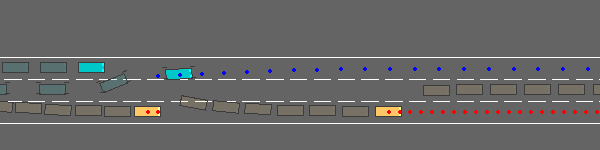}
            \caption[]%
            {{\small \textbf{Phase 4.} During the state "Attempt", the LLM predicts the following vehicle as a cooperative agent and updates the collision time estimation for safety analysis. LLM decides to transit to state "Finish" given all the analysis and the feedback from the reflector and MPC. Finally, it is moving to the target lane safely.}} 
        \end{subfigure}
        \caption[ The average and standard deviation of critical parameters ]
        {\small The ego car is in blue on the middle lane, aiming to cut into the left lane. It interacts with two other blue vehicles in the left lane. The blue dots are the planned trajectory waypoints of the ego. The red dots are the sampled waypoints of other agents from the interval-based prediction. The grey rectangles are the recent trajectory histories of the ego and other agents. The LLM exhibits safe interactive lane-changing behaviors in the multiple-step decision-making process. } 
        \label{fig:interactive_lc}
    \end{figure*}

\noindent
\textbf{Reflection Module:} State and safety checks are stringent requirements in the decision-making process. To ensure compliance with these requirements, a reflection module monitors state transitions. This module corrects and provides feedback to the LLM, facilitating in-context learning, especially when decisions breach these strict constraints. For behavior prediction, the reflection module enforces no constraints to the intention estimation - the LLM independently and flexibly assesses the intentions of surrounding vehicles, categorizing them as either aggressive or cooperative.  

\noindent
\textbf{Intention Prediction Module: } Unlike the prediction for MPC, the intention prediction is to estimate the high-level behavior patterns of the surrounding vehicle, which doesn't need to be very detailed but is important for interaction. We define the potential intention of surrounding agents as cooperative and aggressive. We use the time-to-collision (TTC) of surrounding vehicles as input to the LLM for prediction. At every planning step, the LLM decision-maker will extract the surrounding vehicles' TTCs with the past 3 steps and predict their corresponding intentions. We give several human-labeled demonstrations when setting up the LLM.
\smallskip
\begin{wraptable}{r}{8cm}
\centering
  \caption{Experimental results for lane changing collision rate and success rate with 17 episodes.}
  \begin{tabular}{cccl}
    \toprule
      & Collision Rate & Success Rate &   \\
    \midrule
    DriveLikeAHuman &  47.1\%& 41.2\% &   \\
    \midrule
     Ours &  \textbf{0} & \textbf{100\%} \\
      \midrule
      Ours w/o failsafe &  23.1\% & 76.9\% \\
      \midrule
      Ours w/o reflection&  \textbf{0} & 92.3\% \\
  \bottomrule
\end{tabular}\label{case2_table}
\end{wraptable}
\noindent
\textbf{Experimental Analysis:} In this study, we evaluate our proposed framework using the HighwayEnv simulation platform. As depicted in Figure~\ref{fig:case2_llm_dialogue}, our framework successfully guides the LLM to perform safe motion planning in sequential steps, relying on a state machine, along with prediction and reflection modules.  Figure~\ref{fig:interactive_lc} visualizes the lane-changing scenario, showing the LLM's continuous reasoning and interaction with nearby vehicles under complex conditions. This figure also details the state transitions within the decision-making process. we compared our approach with the open-source DriveLikeAHuman~\cite{fu2023drive} framework in terms of safety (collision rate) and the success rate of lane changes. The findings, presented in Table~\ref{case2_table}, indicate a significantly higher rate of collisions and aborts with the DriveLikeAHuman's naive chain-of-thoughts strategy. In contrast, our method not only ensures safety but also exhibits a remarkable success rate in a variety of generated scenarios, highlighting the efficacy and generalizability of our bi-level interactive planning framework. The final two columns of Table~\ref{case2_table} showcase the significance of our framework's components through an ablation study.

\section{Discussion: LLM as Other Roles for Safety}\label{sec:other_roles}
We discuss the possible usage of LLMs for other components in the AD software pipeline, as shown in Figure~\ref{fig:overview_LLM_AD}. We directly ask  ChatGPT-3.5 (e.g., prompt as ``How can a large language model assist the perception module for safer autonomous driving?'') and summarize its responses below.   

 \textbf{LLMs for Perception.} \textit{1) Multimodal Fusion}: It is possible to consider multimodal infusion with language input.  By integrating information from both sensor data and language input, the perception module can create a more comprehensive understanding of the environment. This multimodal fusion enables the system to make more informed safer decisions by considering both visual information and contextual cues provided by natural language. \textit{2) Semantic Object Recognition}: LLMs can assist in recognizing and understanding objects in the environment based on their semantic context of safety. For instance, if a passenger says, ``Watch out for the cyclist ahead'', LLMs can understand this information to prioritize and adapt the behavior accordingly, enhancing safety. \textit{3) Adaptive Object Detection}: LLMs can provide information that helps the perception module adapt its object detection algorithms based on specific scenarios. For example, if LLMs understand that the vehicle is in a construction zone, they can convey this information to the perception module, prompting the system to be more cautious and attentive to potential hazards. 

\textbf{LLMs for Prediction.} \textit{1) Natural Language Inputs for Contextual Awareness}: The language model in the prediction module can process natural language inputs (possibly from perception) to understand and infer the potential intentions of other drivers. For example, if the perception model or human user interprets "heavy traffic ahead," the prediction module with LLMs can understand it and adjust its expectations and predictions accordingly for safer operation. \textit{2) Human-Centric Predictions}: Language understanding can help the prediction module make more human-centric predictions by considering factors such as hand gestures, turn signals, or spoken commands from other drivers. This allows the autonomous vehicle to anticipate and respond to human behaviors more effectively, improving AV safety.
 \textit{3) Behavioral Evaluation}: The language model can assist in evaluating the driving behaviors and aggressiveness of surrounding cars. This helps the prediction module adjust its predictions based on the perceived driving styles of other vehicles.

\textbf{LLMs for Simulation.}
\textit{1) User Specific Scenario Generation and Variation}: The language model can generate natural language descriptions of diverse driving scenarios by user input for safety concerns, allowing the simulation module to create a wide range of realistic and challenging situations for testing and training in a safety perspective. This helps in ensuring that the autonomous system is well-prepared for various real-world conditions. \textit{2) Human-Like Interaction}: The language model can simulate human-like interactions by generating realistic communication between simulated drivers, pedestrians, and other entities. This enhances the realism of the simulation, allowing the autonomous system to practice responding to natural language cues and gestures for safety purposes. \textit{3) Simulation Annotation and Analysis}: The language model can assist in annotating simulation data by generating descriptions or labels for different events and entities, which further the AV development.  

\section{Conclusion}\label{sec:conclusion}
In conclusion, our presented framework explores the integration of an LLM as an intelligent decision-maker for autonomous driving, fortified by a safety verifier feedback for in-context safety learning.  
Through two case studies, we demonstrate the efficacy of our approach, showcasing notable enhancements in both performance and safety. We further discuss the potential usage of the LLM for other components. This paper intends to broaden the safety perspective within the autonomous driving community concerning the utilization of LLMs. The future directions and remaining challenges include testing this framework in real-world environment and handling ambiguity, biases, and inconsistencies in LLM outputs.

\bibliography{reference}

\begin{thebibliography}{56}
\providecommand{\natexlab}[1]{#1}
\providecommand{\url}[1]{\texttt{#1}}
\expandafter\ifx\csname urlstyle\endcsname\relax
  \providecommand{\doi}[1]{doi: #1}\else
  \providecommand{\doi}{doi: \begingroup \urlstyle{rm}\Url}\fi

\bibitem[Ames et~al.(2019)Ames, Coogan, Egerstedt, Notomista, Sreenath, and Tabuada]{ames2019control}
Aaron~D Ames, Samuel Coogan, Magnus Egerstedt, Gennaro Notomista, Koushil Sreenath, and Paulo Tabuada.
\newblock Control barrier functions: Theory and applications.
\newblock In \emph{2019 18th European control conference (ECC)}, pp.\  3420--3431. IEEE, 2019.

\bibitem[Chen et~al.(2018)Chen, Peng, Grizzle, and Ozay]{chen2018data}
Yuxiao Chen, Huei Peng, Jessy Grizzle, and Necmiye Ozay.
\newblock Data-driven computation of minimal robust control invariant set.
\newblock In \emph{2018 IEEE Conference on Decision and Control (CDC)}, pp.\  4052--4058. IEEE, 2018.

\bibitem[Cui et~al.(2023)Cui, Ma, Cao, Ye, and Wang]{cui2023drive}
Can Cui, Yunsheng Ma, Xu~Cao, Wenqian Ye, and Ziran Wang.
\newblock Drive as you speak: Enabling human-like interaction with large language models in autonomous vehicles.
\newblock \emph{arXiv preprint arXiv:2309.10228}, 2023.

\bibitem[Dawson et~al.(2022)Dawson, Qin, Gao, and Fan]{dawson2022safe}
Charles Dawson, Zengyi Qin, Sicun Gao, and Chuchu Fan.
\newblock Safe nonlinear control using robust neural lyapunov-barrier functions.
\newblock In \emph{Conference on Robot Learning}, pp.\  1724--1735. PMLR, 2022.

\bibitem[Devlin et~al.(2018)Devlin, Chang, Lee, and Toutanova]{devlin2018bert}
Jacob Devlin, Ming-Wei Chang, Kenton Lee, and Kristina Toutanova.
\newblock Bert: Pre-training of deep bidirectional transformers for language understanding.
\newblock \emph{arXiv preprint arXiv:1810.04805}, 2018.

\bibitem[Ding et~al.(2023{\natexlab{a}})Ding, Xu, Arief, Lin, Li, and Zhao]{ding2023survey}
Wenhao Ding, Chejian Xu, Mansur Arief, Haohong Lin, Bo~Li, and Ding Zhao.
\newblock A survey on safety-critical driving scenario generation—a methodological perspective.
\newblock \emph{IEEE Transactions on Intelligent Transportation Systems}, 2023{\natexlab{a}}.

\bibitem[Ding et~al.(2023{\natexlab{b}})Ding, Han, Xu, Zhang, and Li]{ding2023hilm}
Xinpeng Ding, Jianhua Han, Hang Xu, Wei Zhang, and Xiaomeng Li.
\newblock Hilm-d: Towards high-resolution understanding in multimodal large language models for autonomous driving.
\newblock \emph{arXiv preprint arXiv:2309.05186}, 2023{\natexlab{b}}.

\bibitem[Feng et~al.(2020)Feng, Haase-Sch{\"u}tz, Rosenbaum, Hertlein, Glaeser, Timm, Wiesbeck, and Dietmayer]{feng2020deep}
Di~Feng, Christian Haase-Sch{\"u}tz, Lars Rosenbaum, Heinz Hertlein, Claudius Glaeser, Fabian Timm, Werner Wiesbeck, and Klaus Dietmayer.
\newblock Deep multi-modal object detection and semantic segmentation for autonomous driving: Datasets, methods, and challenges.
\newblock \emph{IEEE Transactions on Intelligent Transportation Systems}, 22\penalty0 (3):\penalty0 1341--1360, 2020.

\bibitem[Fu et~al.(2024)Fu, Li, Wen, Dou, Cai, Shi, and Qiao]{fu2023drive}
Daocheng Fu, Xin Li, Licheng Wen, Min Dou, Pinlong Cai, Botian Shi, and Yu~Qiao.
\newblock Drive like a human: Rethinking autonomous driving with large language models.
\newblock In \emph{Proceedings of the IEEE/CVF Winter Conference on Applications of Computer Vision}, pp.\  910--919, 2024.

\bibitem[Gao et~al.(2023)Gao, Chen, Xie, Hong, Li, Yeung, and Xu]{gao2023magicdrive}
Ruiyuan Gao, Kai Chen, Enze Xie, Lanqing Hong, Zhenguo Li, Dit-Yan Yeung, and Qiang Xu.
\newblock Magicdrive: Street view generation with diverse 3d geometry control.
\newblock \emph{arXiv preprint arXiv:2310.02601}, 2023.

\bibitem[Goubault \& Putot(2022)Goubault and Putot]{goubault2022rino}
Eric Goubault and Sylvie Putot.
\newblock Rino: robust inner and outer approximated reachability of neural networks controlled systems.
\newblock In \emph{International Conference on Computer Aided Verification}, pp.\  511--523. Springer, 2022.

\bibitem[Ho et~al.(2020)Ho, Jain, and Abbeel]{NEURIPS2020_4c5bcfec}
Jonathan Ho, Ajay Jain, and Pieter Abbeel.
\newblock Denoising diffusion probabilistic models.
\newblock In H.~Larochelle, M.~Ranzato, R.~Hadsell, M.F. Balcan, and H.~Lin (eds.), \emph{Advances in Neural Information Processing Systems}, volume~33, pp.\  6840--6851. Curran Associates, Inc., 2020.
\newblock URL \url{https://proceedings.neurips.cc/paper_files/paper/2020/file/4c5bcfec8584af0d967f1ab10179ca4b-Paper.pdf}.

\bibitem[Hu et~al.(2023)Hu, Russell, Yeo, Murez, Fedoseev, Kendall, Shotton, and Corrado]{hu2023gaia}
Anthony Hu, Lloyd Russell, Hudson Yeo, Zak Murez, George Fedoseev, Alex Kendall, Jamie Shotton, and Gianluca Corrado.
\newblock Gaia-1: A generative world model for autonomous driving.
\newblock \emph{arXiv preprint arXiv:2309.17080}, 2023.

\bibitem[Huang et~al.(2016)Huang, Chen, Zhang, Qin, Zeng, and Li]{huang2016hierarchical}
Chao Huang, Xin Chen, Yifan Zhang, Shengchao Qin, Yifeng Zeng, and Xuandong Li.
\newblock Hierarchical model predictive control for multi-robot navigation.
\newblock In \emph{Proceedings of the Twenty-Fifth International Joint Conference on Artificial Intelligence}, pp.\  3140--3146, 2016.

\bibitem[Huang et~al.(2019)Huang, Fan, Li, Chen, and Zhu]{huang2019reachnn}
Chao Huang, Jiameng Fan, Wenchao Li, Xin Chen, and Qi~Zhu.
\newblock Reachnn: Reachability analysis of neural-network controlled systems.
\newblock \emph{ACM Transactions on Embedded Computing Systems (TECS)}, 18\penalty0 (5s):\penalty0 1--22, 2019.

\bibitem[Huang et~al.(2022)Huang, Fan, Chen, Li, and Zhu]{huang2022polar}
Chao Huang, Jiameng Fan, Xin Chen, Wenchao Li, and Qi~Zhu.
\newblock Polar: A polynomial arithmetic framework for verifying neural-network controlled systems.
\newblock In \emph{International Symposium on Automated Technology for Verification and Analysis}, pp.\  414--430. Springer, 2022.

\bibitem[Ivanov et~al.(2021)Ivanov, Carpenter, Weimer, Alur, Pappas, and Lee]{ivanov2021verisig}
Radoslav Ivanov, Taylor Carpenter, James Weimer, Rajeev Alur, George Pappas, and Insup Lee.
\newblock Verisig 2.0: Verification of neural network controllers using taylor model preconditioning.
\newblock In \emph{International Conference on Computer Aided Verification}, pp.\  249--262. Springer, 2021.

\bibitem[Jiao et~al.(2022)Jiao, Liu, Zheng, Liang, and Zhu]{jiao2022tae}
Ruochen Jiao, Xiangguo Liu, Bowen Zheng, Dave Liang, and Qi~Zhu.
\newblock Tae: A semi-supervised controllable behavior-aware trajectory generator and predictor.
\newblock In \emph{2022 IEEE/RSJ International Conference on Intelligent Robots and Systems (IROS)}, pp.\  12534--12541. IEEE, 2022.

\bibitem[Jiao et~al.(2023{\natexlab{a}})Jiao, Bai, Liu, Sato, Yuan, Chen, and Zhu]{jiao2023learning}
Ruochen Jiao, Juyang Bai, Xiangguo Liu, Takami Sato, Xiaowei Yuan, Qi~Alfred Chen, and Qi~Zhu.
\newblock Learning representation for anomaly detection of vehicle trajectories.
\newblock In \emph{2023 IEEE/RSJ International Conference on Intelligent Robots and Systems (IROS)}, pp.\  9699--9706. IEEE, 2023{\natexlab{a}}.

\bibitem[Jiao et~al.(2023{\natexlab{b}})Jiao, Liu, Sato, Chen, and Zhu]{jiao2023semi}
Ruochen Jiao, Xiangguo Liu, Takami Sato, Qi~Alfred Chen, and Qi~Zhu.
\newblock Semi-supervised semantics-guided adversarial training for robust trajectory prediction.
\newblock In \emph{Proceedings of the IEEE/CVF International Conference on Computer Vision}, pp.\  8207--8217, 2023{\natexlab{b}}.

\bibitem[Jiao et~al.(2023{\natexlab{c}})Jiao, Wang, Liu, Huang, and Zhu]{jiao2023kinematics}
Ruochen Jiao, Yixuan Wang, Xiangguo Liu, Chao Huang, and Qi~Zhu.
\newblock Kinematics-aware trajectory generation and prediction with latent stochastic differential modeling.
\newblock \emph{arXiv preprint arXiv:2309.09317}, 2023{\natexlab{c}}.

\bibitem[Jin et~al.(2023{\natexlab{a}})Jin, Liu, Zheng, Li, Zhao, Zhang, Zheng, Zhou, and Liu]{jin2023adapt}
Bu~Jin, Xinyu Liu, Yupeng Zheng, Pengfei Li, Hao Zhao, Tong Zhang, Yuhang Zheng, Guyue Zhou, and Jingjing Liu.
\newblock Adapt: Action-aware driving caption transformer.
\newblock \emph{arXiv preprint arXiv:2302.00673}, 2023{\natexlab{a}}.

\bibitem[Jin et~al.(2020)Jin, Wang, Yang, and Mou]{jin2020neural}
Wanxin Jin, Zhaoran Wang, Zhuoran Yang, and Shaoshuai Mou.
\newblock Neural certificates for safe control policies.
\newblock \emph{arXiv preprint arXiv:2006.08465}, 2020.

\bibitem[Jin et~al.(2023{\natexlab{b}})Jin, Shen, Peng, Liu, Qin, Li, Xie, Gao, Zhou, and Gong]{jin2023surrealdriver}
Ye~Jin, Xiaoxi Shen, Huiling Peng, Xiaoan Liu, Jingli Qin, Jiayang Li, Jintao Xie, Peizhong Gao, Guyue Zhou, and Jiangtao Gong.
\newblock Surrealdriver: Designing generative driver agent simulation framework in urban contexts based on large language model.
\newblock \emph{arXiv preprint arXiv:2309.13193}, 2023{\natexlab{b}}.

\bibitem[Kochdumper et~al.(2023)Kochdumper, Krasowski, Wang, Bak, and Althoff]{kochdumper2023provably}
Niklas Kochdumper, Hanna Krasowski, Xiao Wang, Stanley Bak, and Matthias Althoff.
\newblock Provably safe reinforcement learning via action projection using reachability analysis and polynomial zonotopes.
\newblock \emph{IEEE Open Journal of Control Systems}, 2:\penalty0 79--92, 2023.

\bibitem[Leurent(2018)]{highway-env}
Edouard Leurent.
\newblock An environment for autonomous driving decision-making.
\newblock \url{https://github.com/eleurent/highway-env}, 2018.

\bibitem[Li et~al.(2023)Li, Zhang, and Ye]{li2023drivingdiffusion}
Xiaofan Li, Yifu Zhang, and Xiaoqing Ye.
\newblock Drivingdiffusion: Layout-guided multi-view driving scene video generation with latent diffusion model.
\newblock \emph{arXiv preprint arXiv:2310.07771}, 2023.

\bibitem[Liu et~al.(2023{\natexlab{a}})Liu, Hang, Wang, Sun, et~al.]{liu2023mtd}
Jiaqi Liu, Peng Hang, Jianqiang Wang, Jian Sun, et~al.
\newblock Mtd-gpt: A multi-task decision-making gpt model for autonomous driving at unsignalized intersections.
\newblock \emph{arXiv preprint arXiv:2307.16118}, 2023{\natexlab{a}}.

\bibitem[Liu et~al.(2023{\natexlab{b}})Liu, Jiao, Wang, Han, Zheng, and Zhu]{liu2023safety}
Xiangguo Liu, Ruochen Jiao, Yixuan Wang, Yimin Han, Bowen Zheng, and Qi~Zhu.
\newblock Safety-assured speculative planning with adaptive prediction.
\newblock \emph{arXiv preprint arXiv:2307.11876}, 2023{\natexlab{b}}.

\bibitem[Malla et~al.(2023)Malla, Choi, Dwivedi, Choi, and Li]{malla2023drama}
Srikanth Malla, Chiho Choi, Isht Dwivedi, Joon~Hee Choi, and Jiachen Li.
\newblock Drama: Joint risk localization and captioning in driving.
\newblock In \emph{Proceedings of the IEEE/CVF Winter Conference on Applications of Computer Vision}, pp.\  1043--1052, 2023.

\bibitem[Man et~al.(2023)Man, Gui, and Wang]{man2023bev}
Yunze Man, Liang-Yan Gui, and Yu-Xiong Wang.
\newblock Bev-guided multi-modality fusion for driving perception.
\newblock In \emph{Proceedings of the IEEE/CVF Conference on Computer Vision and Pattern Recognition}, pp.\  21960--21969, 2023.

\bibitem[Mao et~al.(2023)Mao, Qian, Zhao, and Wang]{mao2023gpt}
Jiageng Mao, Yuxi Qian, Hang Zhao, and Yue Wang.
\newblock Gpt-driver: Learning to drive with gpt.
\newblock \emph{arXiv preprint arXiv:2310.01415}, 2023.

\bibitem[Nayakanti et~al.(2023)Nayakanti, Al-Rfou, Zhou, Goel, Refaat, and Sapp]{nayakanti2023wayformer}
Nigamaa Nayakanti, Rami Al-Rfou, Aurick Zhou, Kratarth Goel, Khaled~S Refaat, and Benjamin Sapp.
\newblock Wayformer: Motion forecasting via simple \& efficient attention networks.
\newblock In \emph{2023 IEEE International Conference on Robotics and Automation (ICRA)}, pp.\  2980--2987. IEEE, 2023.

\bibitem[OpenAI(2020)]{openai2020chatgpt3}
OpenAI.
\newblock Chatgpt-3: Language model for conversational agents.
\newblock \url{https://www.openai.com/}, 2020.

\bibitem[Prajna(2006)]{prajna2006barrier}
Stephen Prajna.
\newblock Barrier certificates for nonlinear model validation.
\newblock \emph{Automatica}, 42\penalty0 (1):\penalty0 117--126, 2006.

\bibitem[Radford et~al.(2021)Radford, Kim, Hallacy, Ramesh, Goh, Agarwal, Sastry, Askell, Mishkin, Clark, et~al.]{radford2021learning}
Alec Radford, Jong~Wook Kim, Chris Hallacy, Aditya Ramesh, Gabriel Goh, Sandhini Agarwal, Girish Sastry, Amanda Askell, Pamela Mishkin, Jack Clark, et~al.
\newblock Learning transferable visual models from natural language supervision.
\newblock In \emph{International conference on machine learning}, pp.\  8748--8763. PMLR, 2021.

\bibitem[Schilling et~al.(2022)Schilling, Forets, and Guadalupe]{schilling2022verification}
Christian Schilling, Marcelo Forets, and Sebasti{\'a}n Guadalupe.
\newblock Verification of neural-network control systems by integrating taylor models and zonotopes.
\newblock In \emph{Proceedings of the AAAI Conference on Artificial Intelligence}, volume~36, pp.\  8169--8177, 2022.

\bibitem[Sha et~al.(2023)Sha, Mu, Jiang, Chen, Xu, Luo, Li, Tomizuka, Zhan, and Ding]{sha2023languagempc}
Hao Sha, Yao Mu, Yuxuan Jiang, Li~Chen, Chenfeng Xu, Ping Luo, Shengbo~Eben Li, Masayoshi Tomizuka, Wei Zhan, and Mingyu Ding.
\newblock Languagempc: Large language models as decision makers for autonomous driving.
\newblock \emph{arXiv preprint arXiv:2310.03026}, 2023.

\bibitem[Sohl-Dickstein et~al.(2015)Sohl-Dickstein, Weiss, Maheswaranathan, and Ganguli]{sohl2015deep}
Jascha Sohl-Dickstein, Eric Weiss, Niru Maheswaranathan, and Surya Ganguli.
\newblock Deep unsupervised learning using nonequilibrium thermodynamics.
\newblock In \emph{International conference on machine learning}, pp.\  2256--2265. PMLR, 2015.

\bibitem[Touvron et~al.(2023)Touvron, Martin, Stone, Albert, Almahairi, Babaei, Bashlykov, Batra, Bhargava, Bhosale, et~al.]{touvron2023llama}
Hugo Touvron, Louis Martin, Kevin Stone, Peter Albert, Amjad Almahairi, Yasmine Babaei, Nikolay Bashlykov, Soumya Batra, Prajjwal Bhargava, Shruti Bhosale, et~al.
\newblock Llama 2: Open foundation and fine-tuned chat models.
\newblock \emph{arXiv preprint arXiv:2307.09288}, 2023.

\bibitem[Wang et~al.(2023{\natexlab{a}})Wang, Zhu, Huang, Chen, and Lu]{wang2023drivedreamer}
Xiaofeng Wang, Zheng Zhu, Guan Huang, Xinze Chen, and Jiwen Lu.
\newblock Drivedreamer: Towards real-world-driven world models for autonomous driving.
\newblock \emph{arXiv preprint arXiv:2309.09777}, 2023{\natexlab{a}}.

\bibitem[Wang et~al.(2020)Wang, Huang, and Zhu]{wang2020energy}
Yixuan Wang, Chao Huang, and Qi~Zhu.
\newblock Energy-efficient control adaptation with safety guarantees for learning-enabled cyber-physical systems.
\newblock In \emph{Proceedings of the 39th International Conference on Computer-Aided Design}, pp.\  1--9, 2020.

\bibitem[Wang et~al.(2023{\natexlab{b}})Wang, Zhan, Wang, Huang, Wang, Yang, and Zhu]{wang2023joint}
Yixuan Wang, Simon Zhan, Zhilu Wang, Chao Huang, Zhaoran Wang, Zhuoran Yang, and Qi~Zhu.
\newblock Joint differentiable optimization and verification for certified reinforcement learning.
\newblock In \emph{Proceedings of the ACM/IEEE 14th International Conference on Cyber-Physical Systems (with CPS-IoT Week 2023)}, pp.\  132--141, 2023{\natexlab{b}}.

\bibitem[Wang et~al.(2023{\natexlab{c}})Wang, Zhan, Jiao, Wang, Jin, Yang, Wang, Huang, and Zhu]{wang2023enforcing}
Yixuan Wang, Simon~Sinong Zhan, Ruochen Jiao, Zhilu Wang, Wanxin Jin, Zhuoran Yang, Zhaoran Wang, Chao Huang, and Qi~Zhu.
\newblock Enforcing hard constraints with soft barriers: Safe reinforcement learning in unknown stochastic environments.
\newblock In \emph{International Conference on Machine Learning}, pp.\  36593--36604. PMLR, 2023{\natexlab{c}}.

\bibitem[Wang et~al.(2023{\natexlab{d}})Wang, Zhou, Fan, Wang, Li, Chen, Huang, Li, and Zhu]{wang2023polar}
Yixuan Wang, Weichao Zhou, Jiameng Fan, Zhilu Wang, Jiajun Li, Xin Chen, Chao Huang, Wenchao Li, and Qi~Zhu.
\newblock Polar-express: Efficient and precise formal reachability analysis of neural-network controlled systems.
\newblock \emph{IEEE Transactions on Computer-Aided Design of Integrated Circuits and Systems}, 2023{\natexlab{d}}.

\bibitem[Wayve(2023)]{wayve2023lingo}
Wayve.
\newblock Lingo: Natural language for autonomous driving, 2023.
\newblock URL \url{https://wayve.ai/thinking/lingo-natural-language-autonomous-driving/}.
\newblock Accessed: [Insert Access Date].

\bibitem[Wei et~al.(2022)Wei, Wang, Schuurmans, Bosma, Xia, Chi, Le, Zhou, et~al.]{wei2022chain}
Jason Wei, Xuezhi Wang, Dale Schuurmans, Maarten Bosma, Fei Xia, Ed~Chi, Quoc~V Le, Denny Zhou, et~al.
\newblock Chain-of-thought prompting elicits reasoning in large language models.
\newblock \emph{Advances in Neural Information Processing Systems}, 35:\penalty0 24824--24837, 2022.

\bibitem[Wen et~al.(2023)Wen, Fu, Li, Cai, Ma, Cai, Dou, Shi, He, and Qiao]{wen2023dilu}
Licheng Wen, Daocheng Fu, Xin Li, Xinyu Cai, Tao Ma, Pinlong Cai, Min Dou, Botian Shi, Liang He, and Yu~Qiao.
\newblock Dilu: A knowledge-driven approach to autonomous driving with large language models.
\newblock \emph{arXiv preprint arXiv:2309.16292}, 2023.

\bibitem[Wu et~al.(2023)Wu, Han, Wang, Liu, Zhang, and Shen]{wu2023language}
Dongming Wu, Wencheng Han, Tiancai Wang, Yingfei Liu, Xiangyu Zhang, and Jianbing Shen.
\newblock Language prompt for autonomous driving.
\newblock \emph{arXiv preprint arXiv:2309.04379}, 2023.

\bibitem[Xu et~al.(2023)Xu, Zhang, Xie, Zhao, Guo, Wong, Li, and Zhao]{xu2023drivegpt4}
Zhenhua Xu, Yujia Zhang, Enze Xie, Zhen Zhao, Yong Guo, Kenneth~KY Wong, Zhenguo Li, and Hengshuang Zhao.
\newblock Drivegpt4: Interpretable end-to-end autonomous driving via large language model.
\newblock \emph{arXiv preprint arXiv:2310.01412}, 2023.

\bibitem[Yang et~al.(2022)Yang, Chen, Preciado, and Mangharam]{yang2022differentiable}
Shuo Yang, Shaoru Chen, Victor~M Preciado, and Rahul Mangharam.
\newblock Differentiable safe controller design through control barrier functions.
\newblock \emph{IEEE Control Systems Letters}, 7:\penalty0 1207--1212, 2022.

\bibitem[Yang et~al.(2023)Yang, Jia, Li, and Yan]{yang2023survey}
Zhenjie Yang, Xiaosong Jia, Hongyang Li, and Junchi Yan.
\newblock A survey of large language models for autonomous driving.
\newblock \emph{arXiv preprint arXiv:2311.01043}, 2023.

\bibitem[Zhan et~al.(2023)Zhan, Wang, Wu, Jiao, Huang, and Zhu]{zhan2023state}
Simon~Sinong Zhan, Yixuan Wang, Qingyuan Wu, Ruochen Jiao, Chao Huang, and Qi~Zhu.
\newblock State-wise safe reinforcement learning with pixel observations.
\newblock \emph{arXiv preprint arXiv:2311.02227}, 2023.

\bibitem[Zhang et~al.(2023)Zhang, Fu, Zhang, Yu, and Cai]{zhang2023trafficgpt}
Siyao Zhang, Daocheng Fu, Zhao Zhang, Bin Yu, and Pinlong Cai.
\newblock Trafficgpt: Viewing, processing and interacting with traffic foundation models.
\newblock \emph{arXiv preprint arXiv:2309.06719}, 2023.

\bibitem[Zhong et~al.(2023)Zhong, Rempe, Chen, Ivanovic, Cao, Xu, Pavone, and Ray]{zhong2023language}
Ziyuan Zhong, Davis Rempe, Yuxiao Chen, Boris Ivanovic, Yulong Cao, Danfei Xu, Marco Pavone, and Baishakhi Ray.
\newblock Language-guided traffic simulation via scene-level diffusion.
\newblock \emph{arXiv preprint arXiv:2306.06344}, 2023.

\bibitem[Zhou et~al.(2023)Zhou, Liu, Zagar, Yurtsever, and Knoll]{zhou2023vision}
Xingcheng Zhou, Mingyu Liu, Bare~Luka Zagar, Ekim Yurtsever, and Alois~C Knoll.
\newblock Vision language models in autonomous driving and intelligent transportation systems.
\newblock \emph{arXiv preprint arXiv:2310.14414}, 2023.

\end{thebibliography}
\bibliographystyle{iclr2024_conference}


\end{document}